\documentclass{bmvc2k}
\usepackage[utf8]{inputenc}
\usepackage{latexsym}
\usepackage{booktabs}
\usepackage{amsmath,amssymb,amsthm}
\usepackage{todonotes}
\usepackage{array,multirow}
\usepackage{mathtools}
\usepackage{hyperref}
\usepackage{color,soul,colortbl}
\usepackage{tcolorbox}
\usepackage{graphicx}
\usepackage[parfill]{parskip}
\usepackage{xifthen}
\usepackage{diagbox}

\newif\ifsupp
\supptrue

\title{MorphPool: Efficient Non-linear Pooling \& Unpooling in CNNs}

\addauthor{Rick Groenendijk}{r.w.groenendijk@uva.nl}{1}
\addauthor{Leo Dorst}{l.dorst@uva.nl}{1}
\addauthor{Theo Gevers}{th.gevers@uva.nl}{1}

\addinstitution{
 Computer Vision Group\\
 University of Amsterdam \\
 Amsterdam, the Netherlands
}

\runninghead{Groenendijk \etal}{MorphPool}

\def\ie{\emph{i.e}\bmvaOneDot}
\def\eg{\emph{e.g}\bmvaOneDot}

\def\etal{\emph{et al}\bmvaOneDot}

\newcommand{\maxx}[2]{\bigvee\limits_{#1}^{#2}}

\newcommand{\layerout}[1][]{%
  \ifthenelse{\isempty{#1}}%
    {f_{\mathbf{+}}}
    {f_{\mathbf{+}}\left(#1\right)}
}
\newcommand{\layerin}[1][]{%
  \ifthenelse{\isempty{#1}}%
    {f_{\mathbf{-}}}
    {f_{\mathbf{-}}\left(#1\right)}
}
\newcommand{\kernel}[1][]{%
  \ifthenelse{\isempty{#1}}%
    {h}
    {h\left(#1\right)}
}
\newcommand{\layerup}[1][]{%
  \ifthenelse{\isempty{#1}}%
    {f_{\mathbf{\wedge}}}
    {f_{\mathbf{\wedge}}\left(#1\right)}
}
\newcommand{\layerdown}[1][]{%
  \ifthenelse{\isempty{#1}}%
    {f_{\mathbf{\vee}}}
    {f_{\mathbf{\vee}}\left(#1\right)}
}

\newcommand{\ra}[1]{\renewcommand{\arraystretch}{#1}}
\newcolumntype{a}{>{\columncolor{lightgray}}l}

\sethlcolor{gray}

\begin{document}

\maketitle

\begin{abstract}
\noindent Pooling is essentially an operation from the field of Mathematical Morphology, with max pooling as a limited special case.
The more general setting of MorphPooling greatly extends the tool set for building neural networks.
In addition to pooling operations, encoder-decoder networks used for pixel-level predictions also require \emph{unpooling}.
It is common to combine unpooling with convolution or deconvolution for up-sampling.
However, using its morphological properties, unpooling can be generalised and improved.
Extensive experimentation on two tasks and three large-scale datasets shows that morphological pooling and unpooling lead to improved predictive performance at much reduced parameter counts.
\end{abstract}

\section{Introduction} \label{sec:introduction}
Contemporary deep learning architectures exploit pooling operations for two reasons: to filter impactful activation values from feature maps, and to reduce spatial feature size \cite{scherer2010evaluation}.
The most used pooling operation is the \textbf{max pool}, which is used in nearly all common network architectures such as ResNet \cite{he2016deep}, VGGNet \cite{simonyan2014very}, and DenseNet \cite{huang2017densely}.
These network architectures can be applied to pixel-level prediction tasks, such as semantic segmentation.
To do so, inputs are down-sampled to a set of latent features of small spatial size, after which they are up-sampled to full resolution again.
Up-sampling from pooled feature sets most often happens with a combination of unpooling and deconvolution \cite{zeiler2011adaptive,zeiler2014visualizing} and is used in seminal works such as \cite{badrinarayanan2017segnet,noh2015learning,ronneberger2015u}.

As will be shown in this paper, down-sampling using max pooling can be formalised and improved using \emph{mathematical morphology}, the mathematics of contact.
Ever since the works of Serra \cite{serra1982image}, the underlying algebraic structure of data that is acquired using probing contact (\eg LiDAR and radar) has been known to the computer vision community \cite{ritter1996introduction,sussner1998morphological,charisopoulos2017morphological,gartner2008tropical}. 
It is different from the algebra of linear diffusion that is used to build convolutional neural networks (CNNs).
Rather than just considering pooling, it would sensible to formalise pooling \emph{and} unpooling in CNNs as fully morphological operations:
in any encoding layer, pooling is a form of down-sampling and a decoder layer will ultimately need to up-sample.
The task of down-sampling is one of morphological sampling and the task of up-sampling is in fact one of morphological interpolation.

The connection between morphology and pooling has been noted in the literature: 
concurrently with this paper, \cite{angulo2021some} remarks that max pooling and morphological operations indeed share commonalities, although they do not further explore the idea.
\cite{franchi2020deep} proposes a morphological \emph{alternative} to pooling as a weighted combination of morphological dilation and erosion, and calls it morphological pooling.
Both these papers do not formalise the pooling as morphological, nor do they treat up-sampling as the morphological semi-inverse of pooling.
In this article, it is shown how pooling is of an essentially morphological nature (dilation or erosion), and that there is then a naturally associated form of up-sampling (unpooling), dictated by the tropical algebra of morphology.

The contributions of this paper are: \vspace{-0.6em}
\begin{itemize}
    \item A formalisation of max pooling showing that it is an unparameterised non-overlapping special case of the morphological dilation. \vspace{-0.6em}
    \item A fully morphological up-sampling procedure that can replace the semi-morphological unpooling-deconvolution scheme that is commonly used in neural networks. \vspace{-0.6em}
    \item Extensive experimentation to show morphological pooling and unpooling consistently outperform other down and up-sampling schemes at highly reduced parameter counts, including efficient CUDA implementations. This effect is most pronounced on data that is of an inherently geometric nature, such as depth images.
\end{itemize}
\section{Background}

\paragraph{Pooling} \label{sec:background:pooling}
Convolutional networks employ sequential convolution layers to obtain an expressive set of features for regression or classification.
Pooling layers are used to force the network to regress high-quality features, and there exist many pooling schemes (for a review, refer to \cite{gholamalinezhad2020pooling}). 
The most common pooling scheme is the max pool.
Max pooling is used to obtain maximum-amplitude coefficients at intervals while reducing spatial feature sizes \cite{chen2013deep,xie2014spatial}, and to achieve invariance to small distortions from noise \cite{jarrett2009best}.
It is shown to be a very efficient procedure to encode meaningful features, often outperforming mean pooling (\eg \cite{jarrett2009best,yang2009linear}).
In \cite{boureau2010learning}, it is argued that max pooling is more robust to high clutter noise in class separation than mean pooling.

\paragraph{Unpooling} \label{sec:background:unpooling}
While pooling is used to down-sample and refine features, up-sampling may be necessary to obtain predictions at original resolution.
Up-sampling can naturally happen in a variety of ways, but of interest is the inverse of pooling.
\cite{zeiler2011adaptive,zeiler2014visualizing} are the first to define an inverted pooling operation using a combination of deconvolution and 3D reverse max pooling.
Here a set of switches records the location of max values in the pooling step, that are then used to place back latent features at higher resolution;
remaining elements are set to zero.
The unpooling operation is interleaved with deconvolution, or rather transposed convolution, to obtain dense feature maps (\ie infilling).
The main advantage of deconvolution is that it does not employ any predefined interpolation scheme, but can learn the interpolation parameters \footnote{It is however prone to forming checkerboard artefacts \cite{odena2016deconvolution}.}.

There exist many examples in literature of this unpooling-infilling scheme for up-sampling, most notably \cite{noh2015learning,badrinarayanan2017segnet}.
First, \cite{noh2015learning} introduce Deconvolutional Neural Networks.
Their network interleaves 2D unpooling layers with deconvolutions, since it is argued that low-level visual features capture shape details and max pooling has been reported to recover shape well \cite{lee2009convolutional}.
Second, \cite{badrinarayanan2017segnet} introduce SegNet, which interleaves 2D unpooling and convolution. The main motivation is that unpooling results in more sharply delineated semantic boundary predictions.
This strategy is also employed in \cite{long2015fully}, who solve a semantic segmentation task as well.

While \cite{noh2015learning,badrinarayanan2017segnet} employ a deconvolution and a convolution for infilling respectively, it is worth noting that the term deconvolution is misleading: it is not the deconvolution in the mathematical sense, but rather the convolution with flipped kernel (correlation). Since parameters are freely learned in networks (and setting the stride similarly for either operation) they end up being equivalent.

Variations \cite{dosovitskiy2015learning} and improvements \cite{laina2016deeper,xu2019up} of the unpooling-infilling scheme exist, but the (de)convolution operation remains necessary to deal with sparsity.
Whereas pooling is parameterless, the infilling stage is heavily parameterised by (de)convolution to overcome this issue.
It will now be shown that this is not necessary.
\section{Method}
In this section, max pooling will be shown to be a special case of the morphological dilation.
Note that the same could be done for min pooling through erosion, by morphological duality \cite{serra1992overview}.
First, a common notation is introduced for neural networks and individual (linear or morphological) layers within the network.
A neural network $f$ is a composite function $f(\mathbf{x}) = f_{L} \circ \dots \circ f_{l} \circ \dots \circ f_{1} (\mathbf{x})$ of $L$ layers.
Let $\textbf{y}_{n} = f(\mathbf{x}_{n})$ be the output of the network function.
The network is tasked with fitting a (sample, target) dataset $\mathbf{D} = \{ (\mathbf{x}_{n}, \mathbf{t}_{n}) \}_{n=0}^{N}$. 
In this case, training the network is the minimisation of the energy function $E(\mathbf{t}_{n}, \textbf{y}_{n})\, \forall n$, where $\mathbf{t}_{n}$ is the target output.
The input to a specific layer is denoted $\layerin[]$, where the subscript $l$ is dropped for readability. 
The output to that layer is denoted $\layerout[]$, and the parameters of the layer are $\kernel[]$.

\subsection{Morphological Pooling} \label{sec:method:pooling}
Max pooling is a discrete operator that takes a non-overlapping patch-based maximum for any given discrete input map $f(\mathbf{x})$. 
It is shown below that max pooling is a non-overlapping, non-parameterised special case of the morphological dilation.
The morphological dilation is defined on the semi-ring $\left\{\mathbb{R}_{-\infty}, \bigvee, + \right\}$ where $\bigvee$ denotes the supremum operation and $+$ is addition. 
This algebraic system extends the set of real numbers $\mathbb{R}$ with minus infinity: $\mathbb{R}_{-\infty} \equiv \mathbb{R} \cup -\infty$ \cite{ritter1996introduction}.

A layer input signal $\layerin[] \colon \mathbb{R}^{D} \to \mathbb{R}_{-\infty}$ indexed by indicator variable $\mathbf{x}$, and a structuring element $\kernel[] \colon \mathbb{R}^{D} \to \mathbb{R}_{-\infty}$ indexed by indicator variable $\mathbf{z}$ are combined to produce the morphological dilation as the layer output signal $\layerout[]$:

\begin{equation} \label{eq:morphological-dilation}
    \layerout[\mathbf{x}] = \maxx{\mathbf{z}}{} \layerin[\mathbf{x} - \mathbf{z}] + \kernel[\mathbf{z}]\,.
\end{equation}
On the other hand, max pooling can be written as
\begin{equation} \label{eq:max-pool}
    \layerdown[\mathbf{x}] = \mathrm{max}_{\mathbf{z} \in \mathbf{Z}}\left(\layerup[s\mathbf{x} + \mathbf{z}]\right)\,,
\end{equation}
where $\layerdown[]$ is the down-sampled layer output (similar to $\layerout[]$ in~\autoref{eq:morphological-dilation}), $\layerup[]$ is the layer input at high resolution, $s$ is a predefined stride (usually 2), and $\mathbf{z}$ is an indicator set to capture the patch over which the maximum is taken (also usually of size 2). 
The spatial dimensions are thus reduced by stride $s$.

\begin{figure}[t]
\centering
\includegraphics[width=\textwidth]{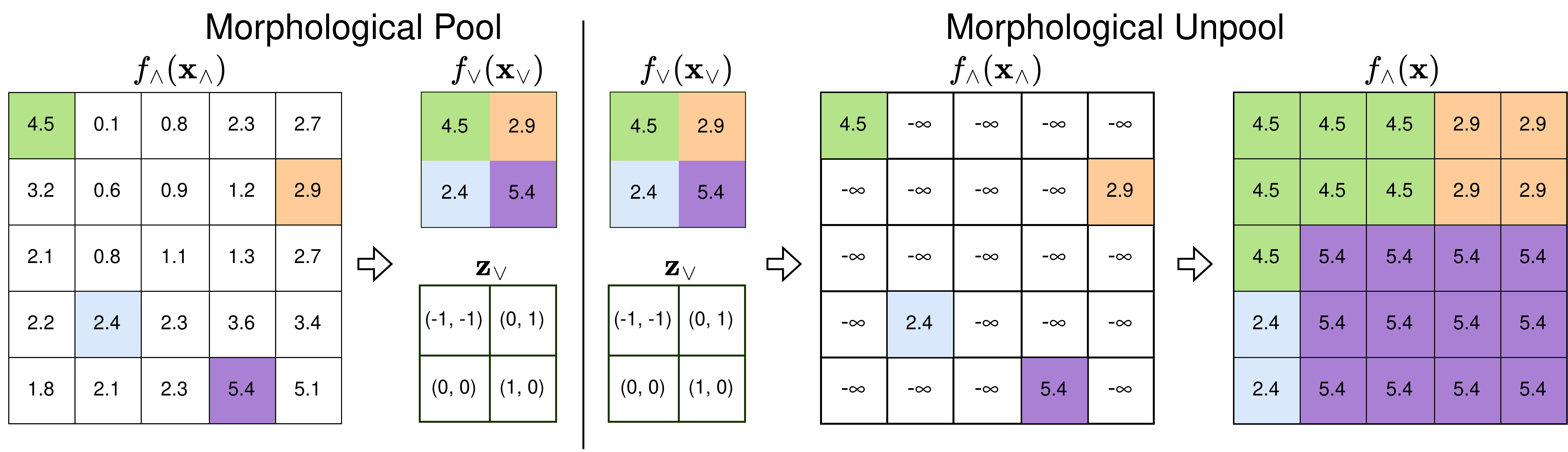}
\caption{
Unparameterised (\ie flat structuring element) morphological pooling \& unpooling.
\textbf{(left)} denotes pooling, which is a dilation with a flat structuring element of size 3x3 combined with sub-sampling resulting in $\layerdown[\mathbf{x_{\vee}}]$ and provenance $\mathbf{z_{\vee}}$. The image is divided in four 3x3 patches;
\textbf{(right)} shows the two-step process of unpooling, which uses the provenance to place the maximum elements at the correct locations resulting in $\layerup[\mathbf{x_{\wedge}}]$; then a 5x5 flat structuring element should be used to morphologically interpolate the maxima yielding the full resolution map $\layerup[\mathbf{x}]$.
}
\label{fig:pool-overview}
\end{figure}

Let $\mathbf{x_{\vee}} \in \mathbf{x}$ be the output locations on the down-sampled output signal $\layerdown[]$. In that case, $\mathbf{x_{\wedge}}$ is the set of input locations scaled by the stride $s$, that is $\left\{\mathbf{x_{\wedge}} | \left(\exists \mathbf{x_{\vee}} \in \mathbf{x}\right) \left[\mathbf{x_{\wedge}} = s\mathbf{x_{\vee}}\right] \right\}$. 
Also, drop the assumption of discreteness to adopt the continuous supremum operator $\bigvee$.
Then, we can rewrite~\autoref{eq:max-pool} to
\begin{equation} \label{eq:max-pool-without-structuring-element}
    \layerdown[\mathbf{x_{\vee}}] = \maxx{\mathbf{z}}{} \layerup[\mathbf{x_{\wedge}} + \mathbf{z}]\,.
\end{equation}
In mathematical morphology, the structuring element $h$ is regarded as a shape that dilates or erodes a signal. 
The simplest structuring element is the \textit{flat} structuring element that can be thought of as a flat disc $\mathbf{Z}$ centred around the origin:
\begin{equation} \label{eq:flat-structuring-element}
    \kernel_{\text{flat}}[\mathbf{z}] = 
    \begin{cases}
    0                   &\text{if}\quad \mathbf{z} \subseteq \mathbf{Z} \\
    -\infty             &\text{otherwise.}
    \end{cases}
\end{equation}
Since $\kernel[]_{\text{flat}}$ is symmetric, the $\mathbf{z}$ indices could be substituted by $-\mathbf{z}$. Thus,~\autoref{eq:max-pool-without-structuring-element} becomes
\begin{equation} \label{eq:formalized-max-pool}
    \layerdown[\mathbf{x_{\vee}}] = \maxx{\mathbf{z}}{} \layerup[\mathbf{x_{\wedge}} - \mathbf{z}] +  \kernel_{\text{flat}}[\mathbf{z}]\,.
\end{equation}
From~\autoref{eq:formalized-max-pool}, it is clear that the max pool is in fact a dilation with a flat structuring element, and the $\vee, \wedge$-notation for the signals $\layerdown[], \layerup[]$ and indicator sets $\mathbf{x_{\vee}}, \mathbf{x_{\wedge}}$ is to account for striding the input by stride $s$.
A schematic overview of pooling by dilation with flat structuring element is given in~\autoref{fig:pool-overview}.

While flat structuring elements are often used in grey-value morphology, there are more interesting structuring elements. 
One example is the parabolic structuring element \cite{boomgaard1996quadratic}:
\begin{equation} \label{eq:parabolic-structuring-element}
    \kernel_{\text{parabolic}}[\mathbf{z}, \sigma] = 
    \begin{cases}
    -\frac{||\mathbf{z}||^{2}}{2\sigma^{2}}         &\text{if}\quad \mathbf{z} \subseteq \mathbf{Z} \\
    -\infty                                         &\text{otherwise.}
    \end{cases}
\end{equation}
Using such a structuring element is a morphological weighting, where pixels closer to the centre affect the maximum more depending on the scale $\sigma$ (and algebraically it is the natural counterpart of Gaussian weighting in convolution \cite{boomgaard1996quadratic}).
Using parabolic $h$, a morphological scale space can be constructed \cite{boomgaard1994morphological,heijmans2002algebraic}.
Scale space is the natural way of dissecting object structure at different scales, and it provides clear intuition for why re-sampling signals can be done by dilation. 
These parabolic structuring elements can for example also be used to build scale equivariant networks \cite{worrall2019deep,sangalli2021scale}.
The flat and parabolic structuring elements are members of a larger set of parameterised morphological operations; the set of parameterised operations generalise max pooling.

Lastly, note that convolution and dilation are logarithmically related in the sense that convolution uses a multiplication-addition scheme where dilation uses an addition-supremum scheme. 
However, convolution on images acts on spatial as well as channel dimensions; grey-value dilation could as well, but pooling is a purely spatial operation.
Therefore, the discussion is limited to purely spatial $h$, omitting channels dimensions from the dilation operation.

\subsection{Morphological Unpooling} \label{sec:method:unpooling}
Morphological unpooling should invert, as much as possible, the pooling operation and up-sample back to original scale. 
Full inversion is not possible, because of non-maximum suppression of the dilation operation.
Morphological unpooling is a two-stage process. 
The first stage is morphological \emph{provenance mapping} \cite{boomgaard1994morphological} (syntactically similar to the morphological derivative introduced in \cite{groenenendijk2022backprop}). 
For purposes of unpooling, $\mathbf{z_{\vee}}$ from the pooling stage can be thought of as provenance: recorded locations at which to place back values during up-sampling.
That is:
\begin{equation}
    \layerup[\mathbf{x_{\wedge}} - \mathbf{z_{\vee}}] =
    \begin{cases}
    \layerdown[\mathbf{x_{\vee}}] &\textup{for all}\quad \mathbf{x_{\vee}} | \left(\exists\mathbf{x_{\wedge}} \in \mathbf{x}\right)\,\left[\mathbf{x_{\wedge}} = s\mathbf{x_{\vee}}\right] \\
    -\infty                           &\textup{otherwise.}
    \end{cases}
\end{equation}
Or, the locations $\mathbf{x_{\vee}}$ on the down-sampled signal $\layerdown[]$ are transferred to the regularly sampled locations $\mathbf{x_{\wedge}}$ compensated by the provenance of the maxima that were stored in $\mathbf{z_{\vee}}$ -- see~\autoref{fig:pool-overview}. 
The values $-\infty$ denote values that are unknown.
This is logarithmically analogous to setting these values to 0 when supplementing this operation with a (de)convolution, because that element would not contribute to the outcome of addition-multiplication.
On the semi-ring $\left\{\mathbb{R}_{-\infty}, \bigvee, + \right\}$, the value $-\infty$ is the 0-element equivalent.

The values set to $-\infty$ at the other locations $\mathbf{x}$, however, cannot be used in training a neural network. 
Since the values $\layerdown[\mathbf{x_{\vee}}]$ at locations $\mathbf{x_{\vee}}$ were maxima in the pooling step, the undefined values at all $\mathbf{x}$ must be upper bounded by $\layerdown[\mathbf{x_{\vee}}]$ and $h$ when unpooling.
Consequently, the second stage is the application of a morphological dilation that retains the same spatial dimensions to morphologically interpolate between the relocated maxima at $\mathbf{x_{\wedge}} - \mathbf{z_{\vee}}$:
\begin{equation} \label{eq:morphological-unpooling}
    \layerup[\mathbf{x}] = \maxx{\mathbf{w}}{} \layerdown[\mathbf{x} - \mathbf{z_{\vee}} - \mathbf{w}] +  \kernel[\mathbf{w}]\,,
\end{equation}
where $\mathbf{w}$ is a scaled indicator set of $s$ times the size of $\mathbf{z}$ to perform gap filling in $\layerup[\mathbf{x_{\wedge}} - \mathbf{z_{\vee}}]$. 
As a result, the output $\layerup[\mathbf{x}]$ is guaranteed not to contain values at $-\infty$; in stead the values have been upper bounded by the maximum $\maxx{\mathbf{w}}{} \layerdown[\mathbf{x} - \mathbf{z_{\vee}} - \mathbf{w}] +  \kernel[\mathbf{w}]$. See~\autoref{fig:pool-overview} for an example with a flat structuring element $h$.

In summary, morphological unpooling is a combination of \textbf{(1)} a strided morphological derivative to up-sample the spatial features and \textbf{(2)} a dilation with increased kernel size to suppress unknown values.
Again, it is possible to parameterise $h$ of the dilation in the second stage (\autoref{eq:morphological-unpooling}) to make full use of concepts from morphological scale space and morphological interpolation.
The full procedure of generalised morphological pooling \& unpooling is called \textbf{MorphPool}.
\section{Experiments}
MorphPool is evaluated on semantic segmentation and depth auto-encoding on NYUv2 \cite{silberman2012nyuv2}, SUN-RGBD \cite{zhou2014learning}, and Stanford 2D-3D-Semantics Dataset (2D-3D-S) \cite{armeni2017joint}.
Morphological unpooling is compared to the standard unpooling-infilling scheme, and to linear sampling and interpolation such as combinations of (de)convolutions.
To this end, DownUpNet is introduced: the simplest encoder-decoder architecture that has variable down and up-sampling blocks. 
The aim of the paper is a proof-of-principle, for which a simple architecture like DownUpNet is sufficient; its goal is not to compete with specialised segmentation networks to achieve state-of-the-art results.
It is hypothesised that morphological operations perform better at depth data, since depth data is inherently non-linear at geometric edges.
In contrast, (de)convolution does not allow for processing 3D spatial information between pixel neighbours, and is sensitive to occlusion.
Implementation details are given below, but more details are found in Supplementary Material \textbf{A}.
Additionally, results in this paper are further supported by Supplementary Material \textbf{B}.

\paragraph{Down \& Up-sampling}
MorphPool is compared to a regular pooling \& unpooling baseline like \eg in \cite{noh2015learning,badrinarayanan2017segnet}.
In those networks, unpooling is followed by (de)convolution to deal with the sparsity that unpooling introduces.
Another baseline is a linear sampling method (common in ResNet \cite{he2016deep} architectures), where down-sampling is a strided convolution and up-sampling is bilinear interpolation followed by a convolution layer.
Both methods introduce high numbers of parameters due to the (de)convolutions.
More specifically, they introduce $C^{2} \times K^{2}$ per layer, where $C$ are the channels and $K$ is the kernel size.
This makes the sampling procedures potentially expressive at the cost of memory.
The morphological pooling \& unpooling with \textbf{flat} structuring elements however, introduce \emph{no} additional parameters.
When the morphological operations are parameterised \textbf{parabolically}, $C$ parameters per layer are introduced.
Finally, for structuring elements that are parameterised at each spatial location --the \textbf{general} setting-- $C \times K^{2}$ parameters per layer are introduced.
It will be shown that additional parameters of (de)convolution are redundant.

\paragraph{DownUpNet}
\begin{figure}[t]
\centering
\includegraphics[width=0.9\textwidth]{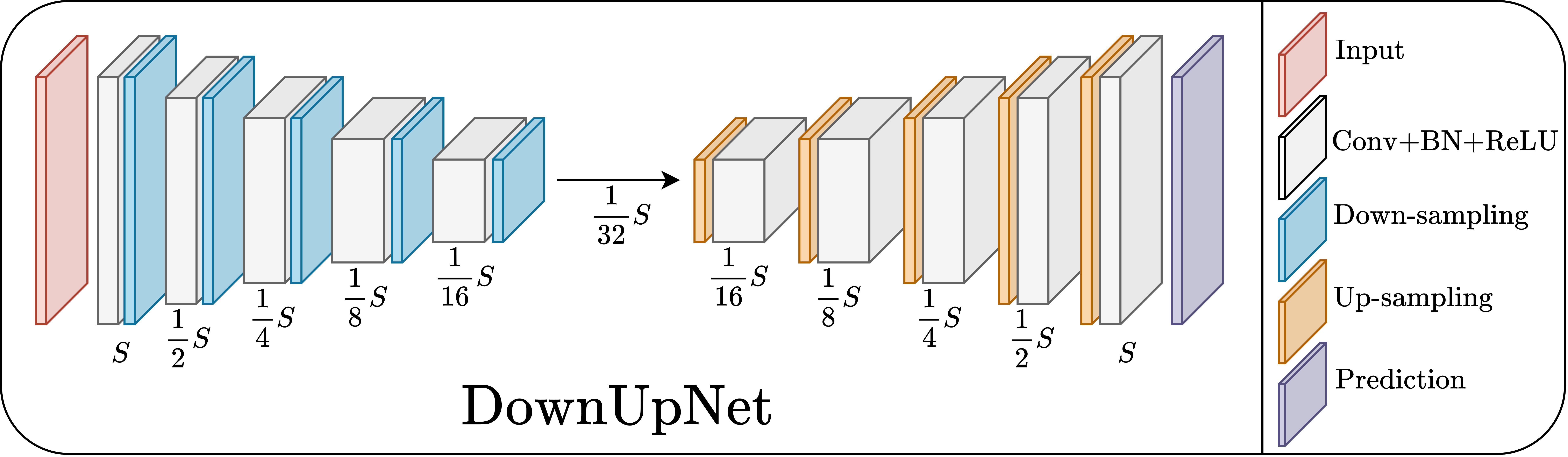}
\caption{
The DownUpNet architecture, which is a simple encoder-decoder.
Each down-sampling operation reduces spatial dimensions $S$ by 2, for a total of $2^{5}$.
The down-sampling operations are preceded by convolution including batch normalisation and ReLU non-linearity.
The up-sampling operations are followed by the same type of block.
}
\label{fig:downupnet-schematic}
\end{figure}
To fairly compare between down and up-sampling schemes, DownUpNet is introduced.
DownUpNet down-samples 5 times, for a total reduction of spatial resolution of $2^{5}$.
It then up-samples back to the original spatial resolution.
Each down-sampling operation is preceded by a convolution, with batch normalisation and ReLU non-linearity.
Each up-sampling operation is followed by a convolution, again with normalisation and non-linearity.
For an overview, see~\autoref{fig:downupnet-schematic}.
Specifically for the larger 2D-3D-S dataset, two convolutions, normalisation, and non-linearities are coupled with sampling operations.

\paragraph{Implementation}
Generalised morphological operations are not implemented by deep learning frameworks such as PyTorch \cite{paszke2019pytorch}.
Implementing the operations in PyTorch as a composed function of additions and maxima is not tractable: unlike convolution, in which multiplication-addition of the patch and the kernel happens jointly, addition-maximum is done in sequence. 
The addition of the kernel to each patch is either memory intensive or time intensive, depending on implementation.
Instead, all morphological operations (\ie forward operations and backward derivatives) in this article are implemented directly in C++/CUDA \cite{nickolls2008scalable}.
More high level functions such as losses and networks are implemented using PyTorch.
All code, including the C++/CUDA code, is made available at \url{https://github.com/rickgroen/morphpool}.

\paragraph{Dataset \& Training}
All experiments are performed on NYUv2 ($N_{\text{train}}=795, N_{\text{test}}=654$), SUN-RGBD ($N_{\text{train}}=5285, N_{\text{test}}=5050$), and 2D-3D-S ($N_{\text{train}}=52903, N_{\text{test}}=17593$).
Experiments are run for both RGB input and for depth input.
RGB images are normalised with training set statistics to follow a zero-mean Gaussian;
depth images are not normalised.
Because of this (and because depth data has more sensor noise and infilling artefacts) networks trained with depth input start with a learning rate $\lambda = 5\mathrm{e}{-4}$ whereas RGB networks start with $\lambda = 5\mathrm{e}{-3}$.
Both learning rates exponentially decay to end at 2\% of the initial $\lambda$.
For training, random crops of size 384$\times$384 are used for NYUv2 and SUN-RGBD, crops of $512\times512$ are used for 2D-3D-S.
During testing, centre crops are used as close to full resolution as possible, while retaining a resolution that is divisible by $2^{5}$.

\subsection{Semantic Segmentation on Depth}
\begin{table*}[t]
    \ra{1.05}
    \centering
    \caption{\textbf{2D-3D-S Segmentation on Depth.}
    The first column denotes the sampling method, the second and third column denote the kernel size for \underline{P}ooling and \underline{U}npooling.
    To deal with issues of sparsity, (de)convolution post-processing of features is necessary; this is denoted at the top as None, Conv and Deconv.
    For each method, the number of \emph{additional} parameters, mean Intersection over Union, pixel accuracy and Boundary F1-score is given.
    The bold-faced results indicate best performance per column.
    Clearly, MorphPool outperforms the alternatives at reduced parameter count.
    }
    \resizebox{1.0\textwidth}{!}{
        \begin{tabular}{@{\extracolsep{4pt}}lcc|cccc|cccc|cccc@{}}
        \toprule
        \multicolumn{3}{c|}{Sampling} & \multicolumn{4}{c}{None} & \multicolumn{4}{c}{Conv} & \multicolumn{4}{c}{Deconv} \\
        &P&U&\#params & mIoU & acc & bf & \#params & mIoU & acc & bf & \#params & mIoU & acc & bf \\
        \cline{1-3} \cline{4-7} \cline{8-11} \cline{12-15} \vspace{-1.0em} \\
        Linear                  & 3 & - &       12.6M & 0.403 & 0.693 & 0.426   & 25.1M & 0.413 & 0.699 & 0.434         & 25.1M & 0.410 & 0.692 & 0.430 \\
        Standard Pool           & 2 & 3 &       0 & 0.352 & 0.657 & 0.375       & 12.6M & 0.377 & 0.675 & 0.393         & 12.6M & 0.397 & 0.693 & 0.407 \\
        \midrule
        MorphPool               & 2 & 3 &       0 & 0.385 & 0.682 & 0.441       & 12.6M & 0.399 & 0.688 & 0.449         & 12.6M & 0.410 & 0.694 & 0.454 \\
        MorphPool               & 3 & 5 &       0 &\bf0.425 &\bf0.707 &\bf0.476 & 12.6M &\bf0.428 &\bf0.714 &\bf0.479   & 12.6M &\bf0.442 &\bf0.720 &\bf0.487 \\
        \bottomrule 
        \end{tabular}
    }
    \label{tab:segmentation-2d3ds-d}
\end{table*}

\begin{table*}[t]
    \ra{1.05}
    \centering
    \caption{\textbf{SUN\&NYU Segmentation on Depth.}
    Additional results on two more datasets, NYUv2 and SUN-RGBD using Depth input.
    MorphPool outperforms the alternatives.
    }
    \resizebox{1.0\textwidth}{!}{
        \begin{tabular}{@{\extracolsep{4pt}}l|ccc|ccc|ccc|ccc@{}}
        \toprule
        &\multicolumn{6}{c}{SUN-RGBD} & \multicolumn{6}{c}{NYU} \\
        \cline{2-7} \cline{8-13}
        & \multicolumn{3}{c}{None} & \multicolumn{3}{c}{Conv} & \multicolumn{3}{c}{None} & \multicolumn{3}{c}{Conv} \\
        & mIoU & acc & bf & mIoU & acc & bf & mIoU & acc & bf & mIoU & acc & bf \\
        \cline{1-1} \cline{2-4} \cline{5-7} \cline{8-10} \cline{11-13} \vspace{-1.0em} \\
        Linear                  & 0.349 & 0.711 & 0.308             & 0.398 & 0.741 & 0.339             & 0.305 & 0.597 & 0.175             & 0.320 & 0.609 & 0.176 \\
        Standard Pool           & 0.208 & 0.643 & 0.255             & 0.297 & 0.687 & 0.294             & 0.121 & 0.453 & 0.177             & 0.159 & 0.495 & 0.173 \\
        MorphPool               & \bf0.382 & \bf0.735 & \bf0.346    & \bf0.412 & \bf0.748 & \bf0.364    & \bf0.323 & \bf0.607 & \bf0.200    & \bf0.357 & \bf0.627 & \bf0.208 \\
        \bottomrule 
        \end{tabular}
    }
    \label{tab:segmentation-nyusun-d}
\end{table*}

First, semantic segmentation experiments are performed on depth input.
Morphological operations are expected to perform well, because depth maps are inherently non-linear in a morphological manner.
That is, morphology allows data to be probed by structuring elements in space; depth is a geometric modality, not a visual one.
Results are shown for 2D-3D-S in~\autoref{tab:segmentation-2d3ds-d}, and for SUN-RGBD and NYUv2 in~\autoref{tab:segmentation-nyusun-d}.
Performance is measured in mean Intersection over Union (mIoU), pixel accuracy, and a boundary F1-score \cite{csurka2004good} to measure performance at semantic edges.
From the results, note: \vspace{-0.6em}
\begin{itemize}
    \item Standard unpooling is outperformed significantly by flat morphological unpooling; introduction of sparse features before (de)convolution hurts performance significantly.\vspace{-0.6em}
    \item It is expected that for a pooling stride of 2, the minimum unpooling with which to fill the features completely is 3. Moreover, with a pooling stride of 3, an unpooling stride of 5 is required. Both are confirmed. \vspace{-0.6em}
    \item Performing strided convolution as down-sampling and bilinear interpolation in combination with convolution for up-sampling does not match performance of morphological sampling on depth data. The addition of two full convolution layers, one for down-sampling and one for up-sampling introduces many parameters to the network. When flat morphological operations are used, there are no additional parameters. \vspace{-0.6em}
    \item For these runs, there is a numerical difference between using convolution and deconvolution. It is expected that these differences fall within a margin due to initialisation; for other experiments, the differences are not so obvious. Convolution and deconvolution should in principle be equivalent. Therefore, only convolution will be used in reporting results from now on. The complete results can be verified in Supp. \textbf{B}. \vspace{-0.6em}
    \item Morphological pooling \& pooling perform best at semantic boundaries, denoted by bf-score. This is in line with the conclusions from \cite{noh2015learning,badrinarayanan2017segnet}, although a fully morphological sampling outperforms those methods. 
\end{itemize}
Besides the results listed above, initial experiments also indicated that mixing linear and non-linear sampling (\eg down-sampling by standard pooling and up-sampling by convolution) had no practical benefits. 
With the natural non-linearities of morphology available, mixing linear and non-linear operations does not seem worthwhile.

\paragraph{Parameterisation of the Structuring Element}
\begin{table*}[t]
    \ra{1.05}
    \centering
    \caption{\textbf{Parameterisation Results.}
    Three different parameterisations for both the morphological pooling \& unpooling are compared, showing that there is a small benefit to using non-flat SEs.
    The SEs introduce several thousands (K) additional parameters.
    }
    \resizebox{1.0\textwidth}{!}{
        \begin{tabular}{@{\extracolsep{4pt}}lc|ccc|ccc|ccc|ccc@{}}
        \toprule
        &&\multicolumn{6}{c}{SUN-RGBD} & \multicolumn{6}{c}{NYU} \\
        \cline{3-8} \cline{9-14}
        && \multicolumn{3}{c}{None} & \multicolumn{3}{c}{Conv} & \multicolumn{3}{c}{None} & \multicolumn{3}{c}{Conv} \\
        &\#params & mIoU & acc & bf & mIoU & acc & bf & mIoU & acc & bf & mIoU & acc & bf \\
        \cline{1-2} \cline{3-5} \cline{6-8} \cline{9-11} \cline{12-14}
        Flat        & 0     & 0.382 & 0.735 & 0.346             & 0.412 & 0.748 & 0.364         & 0.323 & 0.607 & 0.200             & 0.357 & 0.627 & 0.208 \\
        Parabolic   & 4.0K  & 0.396 & 0.741 & 0.351             & 0.412 & \bf0.751 & 0.362      & 0.348 & 0.627 & 0.205             & \bf0.360 & \bf0.630 & \bf0.212 \\
        General     & 67.5K & \bf0.398 & \bf0.745 & \bf0.352    & \bf0.416 & 0.748 & \bf0.366   & \bf0.353 & \bf0.632 & \bf0.207    & 0.357 & \bf0.630 & 0.207 \\
        \bottomrule
        \end{tabular}
    }
    \label{tab:parameterisation-d}
\end{table*}

Besides using a purely flat structuring element (SE) from~\autoref{eq:flat-structuring-element}, it also also possible to parameterise MorphPool.
This could be done using parabolic SEs from~\autoref{eq:parabolic-structuring-element} or by a general parameterised SE: each value on the discrete kernel then has its own parameter that can be freely adjusted.
Results are listed in~\autoref{tab:parameterisation-d} and indicate parameterising the structuring elements is helpful.
General parameterisation of the SEs performs best, although it also introduces most parameters.
Note that compared to introducing an additional convolution layer (\ie millions of parameters), introducing thousands of parameters for the structuring elements is much more computationally reasonable.
General parameterisation of the SEs is used for further experiments.

\begin{table*}
    \ra{1.05}
    \centering
    \caption{\textbf{Depth-wise Convolution Results}. 
    It is possible to use depth-wise convolutions to equalise the number of parameters available during up-sampling.
    Gray cells indicate networks that have the same number of available parameters to learn interpolation for up-sampling, either morphologically or linearly.
    MorphPool outperforms the linear setting.
    }
    \resizebox{1.0\textwidth}{!}{
        \begin{tabular}{@{\extracolsep{4pt}}l|cccc|cccc|cccc@{}}
        \toprule
        \multirow{2}{*}{\diagbox[width=10em,trim=l]{Down}{Up}} & \multicolumn{4}{c}{None} & \multicolumn{4}{c}{Depth-wise Conv} & \multicolumn{4}{c}{Conv}\\
        & \#params & mIoU & acc & bf & \#params & mIoU & acc & bf & \#params & mIoU & acc & bf \\
        \cline{1-1} \cline{2-5} \cline{6-9} \cline{10-13} \vspace{-1.0em} \\
        Conv                    & 12576576 & 0.349 & 0.711 & 0.308    & 12626176 & 0.400 & 0.744 & 0.333     & 47494976 & 0.411 & 0.746 & 0.341 \\
        Depth-wise Conv         & 17856 & 0.356 & 0.716 & 0.313       & \cellcolor{gray!25}67456 & \cellcolor{gray!25}0.370 & \cellcolor{gray!25}0.730 & \cellcolor{gray!25}0.319        & 34936256 & 0.389 & 0.738 & 0.329 \\
        Standard Pool           & 0 & 0.208 & 0.643 & 0.255           & 49600 & 0.285 & 0.693 & 0.285        & 34918400 & 0.333 & 0.720 & 0.313 \\
        \midrule
        MorphPool               & 0 & 0.382 & 0.735 & 0.346     & 49600 & 0.407 & 0.753 & 0.351        & 34918400 & 0.427 & \bf0.757 & \bf0.371 \\
        Para. MorphPool         & 3968 & 0.396 & 0.741 & 0.351     & 53568 & 0.412 & 0.754 & 0.354        & 34922368 & 0.427 & \bf0.757 & 0.366 \\
        General MorphPool       & \cellcolor{gray!25}67456 & \cellcolor{gray!25}\bf0.398 & \cellcolor{gray!25}\bf0.745 & \cellcolor{gray!25}\bf0.352       & 117056 & \bf0.415 & \bf0.755 & \bf0.357       & 34985856 & \bf0.432 & \bf0.757 & \bf0.371 \\
        \bottomrule
        \end{tabular}
    }
    \label{tab:depthwise-results}
\end{table*}
It is possible to compare networks that have the same number of parameters as well.
For this, depth-wise convolutions \cite{chollet2017xception} are used instead of full convolutions; depth-wise convolutions only operate along spatial dimensions.
Results for depth input on the SUN-RGBD dataset are listed in~\autoref{tab:depthwise-results}.
Note especially the gray cells indicating a morphological and linear setting with the same number of parameters.
MorphPool still outperforms the linear setting.

\subsection{Semantic Segmentation on RGB}
\begin{table*}[t]
    \ra{1.05}
    \centering
    \caption{\textbf{Segmentation on RGB.}
    Similar to depth, morphological pooling \& unpooling outperforms standard pooling \& unpooling. 
    Unlike depth, morphological operations are now on par with linear methods.
    This is not surprising given that RGB partly encodes illumination, which is described by linear methods more easily than by morphological methods.
    In contrast, morphological pooling \& unpooling clearly performs best at semantic boundaries as measured by the Boundary F1-score .
    }
    \resizebox{0.85\textwidth}{!}{
        \begin{tabular}{@{\extracolsep{4pt}}l|ccc|ccc|ccc@{}}
        \toprule
        & \multicolumn{3}{c}{2D-3D-S} & \multicolumn{3}{c}{SUN-RGBD} & \multicolumn{3}{c}{NYU} \\
        \cline{2-4} \cline{5-7} \cline{8-10}
        & mIoU & acc & bf & mIoU & acc & bf & mIoU & acc & bf \\
        \cline{1-1} \cline{2-4} \cline{5-7} \cline{8-10}
        Linear                  & \bf0.355 & 0.637 & 0.289          & 0.381 & 0.694 & 0.314             & 0.321 & 0.578 & 0.191 \\
        Standard Pool           & 0.325 & 0.607 & 0.257             & 0.298 & 0.646 & 0.270             & 0.193 & 0.514 & 0.198 \\
        MorphPool               & 0.351 & \bf0.638 & \bf0.310       & \bf0.388 & \bf0.698 & \bf0.337    & \bf0.327 & \bf0.586 & \bf0.211 \\
        \bottomrule
        \end{tabular}
    }
    \label{tab:segmentation-2d3dsnyusun-rgb}
\end{table*}

Semantic segmentation is also performed on RGB input (see~\autoref{tab:segmentation-2d3dsnyusun-rgb}).
It can be concluded that, again, regular pooling and unpooling yields worse performance than generalised morphological pooling and unpooling on semantic segmentation.
Now, convolution and interpolation perform on par with morphology.
Note, however, that the comparison is not completely fair due to the increased number of parameters convolution has available.
Interpolation networks have 50\% more parameters than morphological methods.
Like for semantic segmentation using depth, morphological pooling \& pooling for RGB input perform best at semantic boundaries.
Performance of morphological pooling and unpooling on depth-based segmentation outperforms RGB-based segmentation.
Compared to depth, it is understandable that RGB signals are less easily encoded using mathematical morphology: RGB does not just express geometry but also illumination, which has aspects that are well described linearly.

\subsection{Depth Auto-encoding}
\begin{table*}[t]
    \ra{1.05}
    \centering
    \caption{\textbf{Depth Auto-encoding Results.}
    Similar to segmentation on depth, morphological operations are well suited to sampling features from a depth modality on an auto-encoding task.
    MorphPool outperforms linear sampling and standard pooling on this task.
    }
    \resizebox{0.8\textwidth}{!}{
        \begin{tabular}{@{\extracolsep{4pt}}l|ccc|ccc@{}}
        \toprule
        & \multicolumn{3}{c}{None} & \multicolumn{3}{c}{Conv} \\
        & ARD ($\downarrow$) & RMS ($\downarrow$) & $\delta_{t}<$ 1.25 ($\uparrow$) & ARD ($\downarrow$) & RMS ($\downarrow$) & $\delta_{t}<$ 1.25 ($\uparrow$) \\
        \cline{1-1} \cline{2-4} \cline{5-7}
        Linear                  & 0.193 & 0.563 & 0.715                 & 0.190 & 0.566 & 0.716 \\
        Standard Pool           & 0.239 & 0.751 & 0.605                 & 0.224 & 0.687 & 0.634 \\
        MorphPool               & \bf0.171 & \bf0.505 & \bf0.747        & \bf0.161 & \bf0.462 & \bf0.774 \\
        \bottomrule
        \end{tabular}
    }
    \label{tab:encoding-results}
\end{table*}

As a final experiment, a depth auto-encoding task is solved; results are shown in~\autoref{tab:encoding-results}.
Performance is measured in Absolute Relative Distance (ARD), Root Mean Squared Error (RMS), and an accuracy $\delta_{t}$ of predictions within a threshold $t=1.25$.
The results show substantial improvement, confirming depth is a modality that is very well suited to the use of morphological operations.

\section{Conclusions}
In this paper, max pooling has been formalised as a non-parameterised non-overlapping special case of the more general morphological dilation.
In view of this, the unpooling-infilling scheme can be replaced by a generalised morphological operation. 
The full procedure is called \textbf{MorphPool}.
Experiments on two tasks and three datasets show that the proposed method outperforms other sampling schemes at much reduced parameter counts. 
The effect is most pronounced on depth data, which confirms the expectation that \emph{mathematical morphology} may be the natural language to process such non-linear geometry.
Given the success of morphological operations for processing depth, future research could investigate whether fully morphological encoders may be used in multi-modal networks. 
In that case, parallel treatment of linear and morphological features has to explored.

\paragraph{Acknowledgements}
This publication is part of the project FlexCRAFT (with project number P17-01) which is (partly) financed by the Dutch Research Council (NWO).

\bibliography{refs}

\begin{thebibliography}{43}
\providecommand{\natexlab}[1]{#1}
\providecommand{\url}[1]{\texttt{#1}}
\expandafter\ifx\csname urlstyle\endcsname\relax
  \providecommand{\doi}[1]{doi: #1}\else
  \providecommand{\doi}{doi: \begingroup \urlstyle{rm}\Url}\fi

\bibitem[Angulo(2021)]{angulo2021some}
Jesus Angulo.
\newblock Some open questions on morphological operators and representations in
  the deep learning era.
\newblock In \emph{International Conference on Discrete Geometry and
  Mathematical Morphology}, pages 3--19. Springer, 2021.

\bibitem[Armeni et~al.(2017)Armeni, Sax, Zamir, and Savarese]{armeni2017joint}
Iro Armeni, Sasha Sax, Amir~R Zamir, and Silvio Savarese.
\newblock Joint 2d-3d-semantic data for indoor scene understanding.
\newblock \emph{arXiv preprint arXiv:1702.01105}, 2017.

\bibitem[Badrinarayanan et~al.(2017)Badrinarayanan, Kendall, and
  Cipolla]{badrinarayanan2017segnet}
Vijay Badrinarayanan, Alex Kendall, and Roberto Cipolla.
\newblock Segnet: A deep convolutional encoder-decoder architecture for image
  segmentation.
\newblock \emph{IEEE transactions on pattern analysis and machine
  intelligence}, 39\penalty0 (12):\penalty0 2481--2495, 2017.

\bibitem[Boureau et~al.(2010)Boureau, Bach, LeCun, and
  Ponce]{boureau2010learning}
Y-Lan Boureau, Francis Bach, Yann LeCun, and Jean Ponce.
\newblock Learning mid-level features for recognition.
\newblock In \emph{2010 IEEE computer society conference on computer vision and
  pattern recognition}, pages 2559--2566. IEEE, 2010.

\bibitem[Charisopoulos and Maragos(2017)]{charisopoulos2017morphological}
Vasileios Charisopoulos and Petros Maragos.
\newblock Morphological perceptrons: geometry and training algorithms.
\newblock In \emph{International Symposium on Mathematical Morphology and Its
  Applications to Signal and Image Processing}, pages 3--15. Springer, 2017.

\bibitem[Chen et~al.(2013)Chen, Polatkan, Sapiro, Blei, Dunson, and
  Carin]{chen2013deep}
Bo~Chen, Gungor Polatkan, Guillermo Sapiro, David Blei, David Dunson, and
  Lawrence Carin.
\newblock Deep learning with hierarchical convolutional factor analysis.
\newblock \emph{IEEE transactions on pattern analysis and machine
  intelligence}, 35\penalty0 (8):\penalty0 1887--1901, 2013.

\bibitem[Chollet(2017)]{chollet2017xception}
Fran{\c{c}}ois Chollet.
\newblock Xception: Deep learning with depthwise separable convolutions.
\newblock In \emph{Proceedings of the IEEE conference on computer vision and
  pattern recognition}, pages 1251--1258, 2017.

\bibitem[Csurka et~al.(2004)Csurka, Larlus, Perronnin, and
  Meylan]{csurka2004good}
Gabriela Csurka, Diane Larlus, Florent Perronnin, and France Meylan.
\newblock What is a good evaluation measure for semantic segmentation?
\newblock \emph{IEEE PAMI}, 26\penalty0 (1), 2004.

\bibitem[Dosovitskiy et~al.(2015)Dosovitskiy, Tobias~Springenberg, and
  Brox]{dosovitskiy2015learning}
Alexey Dosovitskiy, Jost Tobias~Springenberg, and Thomas Brox.
\newblock Learning to generate chairs with convolutional neural networks.
\newblock In \emph{Proceedings of the IEEE conference on computer vision and
  pattern recognition}, pages 1538--1546, 2015.

\bibitem[Franchi et~al.(2020)Franchi, Fehri, and Yao]{franchi2020deep}
Gianni Franchi, Amin Fehri, and Angela Yao.
\newblock Deep morphological networks.
\newblock \emph{Pattern Recognition}, 102:\penalty0 107246, 2020.

\bibitem[G{\"a}rtner and Jaggi(2008)]{gartner2008tropical}
Bernd G{\"a}rtner and Martin Jaggi.
\newblock Tropical support vector machines.
\newblock Technical report, Citeseer, 2008.

\bibitem[Gholamalinezhad and Khosravi(2020)]{gholamalinezhad2020pooling}
Hossein Gholamalinezhad and Hossein Khosravi.
\newblock Pooling methods in deep neural networks, a review.
\newblock \emph{arXiv preprint arXiv:2009.07485}, 2020.

\bibitem[Groenendijk et~al.(2022)Groenendijk, Dorst, and
  Gevers]{groenenendijk2022backprop}
Rick Groenendijk, Leo Dorst, and Theo Gevers.
\newblock Geometric back-propagation in morphological neural networks.
\newblock \emph{TechRxiv preprint}, 2022.
\newblock \doi{https://doi.org/10.36227/techrxiv.20330667.v1}.

\bibitem[He et~al.(2016)He, Zhang, Ren, and Sun]{he2016deep}
Kaiming He, Xiangyu Zhang, Shaoqing Ren, and Jian Sun.
\newblock Deep residual learning for image recognition.
\newblock In \emph{Proceedings of the IEEE conference on computer vision and
  pattern recognition}, pages 770--778, 2016.

\bibitem[Heijmans and van~den Boomgaard(2002)]{heijmans2002algebraic}
Henk Heijmans and Rein van~den Boomgaard.
\newblock Algebraic framework for linear and morphological scale-spaces.
\newblock \emph{Journal of Visual Communication and Image Representation},
  13\penalty0 (1-2):\penalty0 269--301, 2002.

\bibitem[Huang et~al.(2017)Huang, Liu, Van Der~Maaten, and
  Weinberger]{huang2017densely}
Gao Huang, Zhuang Liu, Laurens Van Der~Maaten, and Kilian~Q Weinberger.
\newblock Densely connected convolutional networks.
\newblock In \emph{Proceedings of the IEEE conference on computer vision and
  pattern recognition}, pages 4700--4708, 2017.

\bibitem[Jarrett et~al.(2009)Jarrett, Kavukcuoglu, Ranzato, and
  LeCun]{jarrett2009best}
Kevin Jarrett, Koray Kavukcuoglu, Marc'Aurelio Ranzato, and Yann LeCun.
\newblock What is the best multi-stage architecture for object recognition?
\newblock In \emph{2009 IEEE 12th international conference on computer vision},
  pages 2146--2153. IEEE, 2009.

\bibitem[Laina et~al.(2016)Laina, Rupprecht, Belagiannis, Tombari, and
  Navab]{laina2016deeper}
Iro Laina, Christian Rupprecht, Vasileios Belagiannis, Federico Tombari, and
  Nassir Navab.
\newblock Deeper depth prediction with fully convolutional residual networks.
\newblock In \emph{2016 Fourth international conference on 3D vision (3DV)},
  pages 239--248. IEEE, 2016.

\bibitem[Lee et~al.(2009)Lee, Grosse, Ranganath, and Ng]{lee2009convolutional}
Honglak Lee, Roger Grosse, Rajesh Ranganath, and Andrew~Y Ng.
\newblock Convolutional deep belief networks for scalable unsupervised learning
  of hierarchical representations.
\newblock In \emph{Proceedings of the 26th annual international conference on
  machine learning}, pages 609--616, 2009.

\bibitem[Long et~al.(2015)Long, Shelhamer, and Darrell]{long2015fully}
Jonathan Long, Evan Shelhamer, and Trevor Darrell.
\newblock Fully convolutional networks for semantic segmentation.
\newblock In \emph{Proceedings of the IEEE conference on computer vision and
  pattern recognition}, pages 3431--3440, 2015.

\bibitem[Nickolls et~al.(2008)Nickolls, Buck, Garland, and
  Skadron]{nickolls2008scalable}
John Nickolls, Ian Buck, Michael Garland, and Kevin Skadron.
\newblock Scalable parallel programming with cuda: Is cuda the parallel
  programming model that application developers have been waiting for?
\newblock \emph{Queue}, 6\penalty0 (2):\penalty0 40--53, 2008.

\bibitem[Noh et~al.(2015)Noh, Hong, and Han]{noh2015learning}
Hyeonwoo Noh, Seunghoon Hong, and Bohyung Han.
\newblock Learning deconvolution network for semantic segmentation.
\newblock In \emph{Proceedings of the IEEE international conference on computer
  vision}, pages 1520--1528, 2015.

\bibitem[Odena et~al.(2016)Odena, Dumoulin, and Olah]{odena2016deconvolution}
Augustus Odena, Vincent Dumoulin, and Chris Olah.
\newblock Deconvolution and checkerboard artifacts.
\newblock \emph{Distill}, 2016.
\newblock \doi{10.23915/distill.00003}.
\newblock URL \url{http://distill.pub/2016/deconv-checkerboard}.

\bibitem[Paszke et~al.(2019)Paszke, Gross, Massa, Lerer, Bradbury, Chanan,
  Killeen, Lin, Gimelshein, Antiga, Desmaison, Kopf, Yang, DeVito, Raison,
  Tejani, Chilamkurthy, Steiner, Fang, Bai, and Chintala]{paszke2019pytorch}
Adam Paszke, Sam Gross, Francisco Massa, Adam Lerer, James Bradbury, Gregory
  Chanan, Trevor Killeen, Zeming Lin, Natalia Gimelshein, Luca Antiga, Alban
  Desmaison, Andreas Kopf, Edward Yang, Zachary DeVito, Martin Raison, Alykhan
  Tejani, Sasank Chilamkurthy, Benoit Steiner, Lu~Fang, Junjie Bai, and Soumith
  Chintala.
\newblock Pytorch: An imperative style, high-performance deep learning library.
\newblock In \emph{Advances in Neural Information Processing Systems 32}, pages
  8024--8035. Curran Associates, Inc., 2019.

\bibitem[Ritter and Sussner(1996)]{ritter1996introduction}
Gerhard~X Ritter and Peter Sussner.
\newblock An introduction to morphological neural networks.
\newblock In \emph{Proceedings of 13th International Conference on Pattern
  Recognition}, volume~4, pages 709--717. IEEE, 1996.

\bibitem[Ronneberger et~al.(2015)Ronneberger, Fischer, and
  Brox]{ronneberger2015u}
Olaf Ronneberger, Philipp Fischer, and Thomas Brox.
\newblock U-net: Convolutional networks for biomedical image segmentation.
\newblock In \emph{International Conference on Medical image computing and
  computer-assisted intervention}, pages 234--241. Springer, 2015.

\bibitem[Sangalli et~al.(2021)Sangalli, Blusseau, Velasco-Forero, and
  Angulo]{sangalli2021scale}
Mateus Sangalli, Samy Blusseau, Santiago Velasco-Forero, and Jesus Angulo.
\newblock Scale equivariant neural networks with morphological scale-spaces.
\newblock In \emph{International Conference on Discrete Geometry and
  Mathematical Morphology}, pages 483--495. Springer, 2021.

\bibitem[Scherer et~al.(2010)Scherer, M{\"u}ller, and
  Behnke]{scherer2010evaluation}
Dominik Scherer, Andreas M{\"u}ller, and Sven Behnke.
\newblock Evaluation of pooling operations in convolutional architectures for
  object recognition.
\newblock In \emph{International conference on artificial neural networks},
  pages 92--101. Springer, 2010.

\bibitem[Serra(1983)]{serra1982image}
Jean Serra.
\newblock Image analysis and mathematical morphology.
\newblock 1983.

\bibitem[Serra and Vincent(1992)]{serra1992overview}
Jean Serra and Luc Vincent.
\newblock An overview of morphological filtering.
\newblock \emph{Circuits, Systems and Signal Processing}, 11\penalty0
  (1):\penalty0 47--108, 1992.

\bibitem[Silberman et~al.(2012)Silberman, Hoiem, Kohli, and
  Fergus]{silberman2012nyuv2}
Nathan Silberman, Derek Hoiem, Pushmeet Kohli, and Rob Fergus.
\newblock Indoor segmentation and support inference from rgbd images.
\newblock In \emph{ECCV}, 2012.

\bibitem[Simonyan and Zisserman(2014)]{simonyan2014very}
Karen Simonyan and Andrew Zisserman.
\newblock Very deep convolutional networks for large-scale image recognition.
\newblock \emph{arXiv preprint arXiv:1409.1556}, 2014.

\bibitem[Sussner(1998)]{sussner1998morphological}
Peter Sussner.
\newblock Morphological perceptron learning.
\newblock In \emph{Proceedings of the 1998 ISIC held jointly with CIRA}, pages
  477--482. IEEE, 1998.

\bibitem[Sutskever et~al.(2013)Sutskever, Martens, Dahl, and
  Hinton]{sutskever2013importance}
Ilya Sutskever, James Martens, George Dahl, and Geoffrey Hinton.
\newblock On the importance of initialization and momentum in deep learning.
\newblock In \emph{International conference on machine learning}, pages
  1139--1147. PMLR, 2013.

\bibitem[van~den Boomgaard and Smeulders(1994)]{boomgaard1994morphological}
Rein van~den Boomgaard and Arnold Smeulders.
\newblock The morphological structure of images: The differential equations of
  morphological scale-space.
\newblock \emph{IEEE transactions on pattern analysis and machine
  intelligence}, 16\penalty0 (11):\penalty0 1101--1113, 1994.

\bibitem[van~den Boomgaard et~al.(1996)van~den Boomgaard, Dorst, Makram-Ebeid,
  and Schavemaker]{boomgaard1996quadratic}
Rein van~den Boomgaard, Leo Dorst, Sherif Makram-Ebeid, and John Schavemaker.
\newblock Quadratic structuring functions in mathematical morphology.
\newblock In \emph{Mathematical morphology and its applications to image and
  signal processing}, pages 147--154. Springer, 1996.

\bibitem[Worrall and Welling(2019)]{worrall2019deep}
Daniel Worrall and Max Welling.
\newblock Deep scale-spaces: Equivariance over scale.
\newblock \emph{Advances in Neural Information Processing Systems}, 32, 2019.

\bibitem[Xie et~al.(2014)Xie, Tian, Wang, and Zhang]{xie2014spatial}
Lingxi Xie, Qi~Tian, Meng Wang, and Bo~Zhang.
\newblock Spatial pooling of heterogeneous features for image classification.
\newblock \emph{IEEE Transactions on Image Processing}, 23\penalty0
  (5):\penalty0 1994--2008, 2014.

\bibitem[Xu et~al.(2019)Xu, Yang, Lai, Gao, Shen, and Yan]{xu2019up}
Chunyan Xu, Jian Yang, Hanjiang Lai, Junbin Gao, Linlin Shen, and Shuicheng
  Yan.
\newblock Up-cnn: Un-pooling augmented convolutional neural network.
\newblock \emph{Pattern Recognition Letters}, 119:\penalty0 34--40, 2019.

\bibitem[Yang et~al.(2009)Yang, Yu, Gong, and Huang]{yang2009linear}
Jianchao Yang, Kai Yu, Yihong Gong, and Thomas Huang.
\newblock Linear spatial pyramid matching using sparse coding for image
  classification.
\newblock In \emph{2009 IEEE Conference on computer vision and pattern
  recognition}, pages 1794--1801. IEEE, 2009.

\bibitem[Zeiler and Fergus(2014)]{zeiler2014visualizing}
Matthew~D Zeiler and Rob Fergus.
\newblock Visualizing and understanding convolutional networks.
\newblock In \emph{European conference on computer vision}, pages 818--833.
  Springer, 2014.

\bibitem[Zeiler et~al.(2011)Zeiler, Taylor, and Fergus]{zeiler2011adaptive}
Matthew~D Zeiler, Graham~W Taylor, and Rob Fergus.
\newblock Adaptive deconvolutional networks for mid and high level feature
  learning.
\newblock In \emph{2011 international conference on computer vision}, pages
  2018--2025. IEEE, 2011.

\bibitem[Zhou et~al.(2014)Zhou, Lapedriza, Xiao, Torralba, and
  Oliva]{zhou2014learning}
Bolei Zhou, Agata Lapedriza, Jianxiong Xiao, Antonio Torralba, and Aude Oliva.
\newblock Learning deep features for scene recognition using places database.
\newblock \emph{Advances in neural information processing systems}, 27, 2014.

\end{thebibliography}

\ifsupp
\newpage
\section*{Supplementary Material A: Implementation Details}

\subsection*{Network Architecture}
Using the DownUpNet architecture, the quality of semantic predictions as a result of different sampling schemes is tested.
A schematic overview of the network architecture is shown in Figure 2.
There are three choices for down-sampling operations: \vspace{-0.6em}
\begin{itemize}
    \item Convolution with stride 2 and kernel size 3. This operations introduces $C^{2} \times K^{2}$ additional parameters per down-sampling layer, where $C$ is the number of channels of the layer and $K$ the kernel size. \vspace{-0.6em}
    \item Max Pool with a non-overlapping kernel of size 2 at stride 2. \vspace{-0.6em}
    \item Generalised Morphological Pooling, with variable kernel size. The pooling operation can be flat (similar to max pooling) or be parameterized with any structuring element. In the case of parabolic structuring elements, this layer introduces $C$ parameters. And in the freely parameterized case, this introduced $C \times K^{2}$ parameters.
\end{itemize}
There are three choices for up-sampling operations: \vspace{-0.6em}
\begin{itemize}
    \item Sparse up-sampling at rate 2, followed by bilinear interpolation and convolution. This operation introduces $C^{2} \times K^{2}$ additional parameters per up-sampling layer. \vspace{-0.6em}
    \item Unpooling followed by deconvolution as introduced in \cite{zeiler2011adaptive} and used in \cite{noh2015learning}. This operation introduces $C^{2} \times K^{2}$ additional parameters per up-sampling layer. \vspace{-0.6em}
    \item Morphological unpooling as described in Section 3.2, which unpools by provenance followed by a morphological dilation. In case of a flat structuring element, this is a parameterless layer. In case of parabolic parameterisation, this introduces $C$ parameters. And in the case of a freely parameterized dilation, this layer introduces $C \times K^{2}$ additional parameters.
\end{itemize}
Note especially that morphological pooling and unpooling (\ie MorphPool) introduce hardly any additional parameters. Even the general (\ie freely) parameterised dilations introduce a factor $C$ (\ie the number of channels) fewer parameters than full convolutions in the other sampling operations do.

In the network, all convolutions have a kernel size of 3, and are padded to retain resolution.
When training on 2D-3D-S, there are two convolution blocks (including batch normalisation and ReLU non-linearity) before and after each sampling operation.
This is because the 2D-3D-S is much larger than NYUv2 and SUN-RGBD and initial experimentation showed significantly improved performance of this architecture using a larger network for all methods.

\subsection*{Training details}
All methods are implemented in PyTorch, with the exception of the morphological operations which are implemented as C++/CUDA extensions.
The extensions are compiled using g++ and CUDA11.
All networks are trained using SGD with Nesterov Momentum \cite{sutskever2013importance}, with a learning rate that decays exponentially by a scalar $\gamma$.
Based on initial experimentation, for consistent performance this $\gamma$ can be set such that the initial learning rate $\lambda$ decays to  at 2\% of the initial $\lambda$, according to $\gamma = \sqrt[\text{epochs}]{\frac{2}{100}}$ with \textit{epochs} the number of training epochs, and $\textit{epochs} \geq 1$.

RGB images are normalised with training set statistics to follow a zero-mean Gaussian, as is common for neural networks.
Depth images however are not normalised, and are unscaled depth values in meters.
Depth data in general has more sensor noise and infilling artefacts.
In combination with the unscaled data, this led to less stable learning at similar $\lambda$ to RGB images.
Therefore, networks trained with depth input start with a learning rat $\lambda = 5\mathrm{e}{-4}$, whereas RGB input starts with $\lambda = 5\mathrm{e}{-3}$.

During training, random crops of size 384$\times$384 are used for NYUv2 and SUN-RGBD, crops of $512\times512$ are used for 2D-3D-S.
During testing, centre crops are used as close to full resolution as possible, while retaining a resolution that is divisible by $2^{5}$.
Networks are trained for 100 epochs on NYUv2 and SUN-RGBD with a batch size of 8; on 2D-3D-S networks are trained for 40 epochs with a batch size of 16. 
The experiments can be run on a NVIDIA GTX1080Ti, except the experiments on 2D-3D-S which is trained on a NVIDIA A6000.
\section*{Supplementary Material B: Supporting Results}
In this section, all supporting results for the main experiments section are given.
All results are reported over a variety of tables.
For semantic segmentation on Depth, see~\autoref{tab:full-segmentation-2d3ds-d} for 2D-3D-S, \autoref{tab:full-segmentation-sun-d} for SUN-RGBD, and \autoref{tab:full-segmentation-nyu-d} for NYUv2.
For semantic segmentation on RGB, see~\autoref{tab:full-segmentation-2d3ds-i} for 2D-3D-S, \autoref{tab:full-segmentation-sun-i} for SUN-RGBD, and \autoref{tab:full-segmentation-nyu-i} for NYUv2.
And finally, for depth auto-encoding, see~\autoref{tab:full-encoding-nyu}.

\begin{table*}
    \ra{1.05}
    \centering
    \caption{\textbf{2D-3D-S Segmentation on Depth Results.}
    The first column denotes the sampling method, the second through fourth column denotes the kernel size with which the \underline{P}ooling, \underline{U}npooling, and \underline{C}onvolution scheme worked. 
    For each method of post-processing (None, Convolution, Deconvolution) the up-sampled feature volume, the number of parameters and performance is given.
    The high-lighted gray cells denote the standard procedures for up and down-sampling: Unpooling followed by (de)convolution, bilinear interpolation followed by convolution.
    In addition, General MorphPool denotes general parameterised morphological structuring elements.
    The bold-faced results indicate best performance per column.
    }
    \resizebox{1.0\textwidth}{!}{
        \begin{tabular}{@{\extracolsep{4pt}}lccc|cccc|cccc|cccc@{}}
        \toprule
        &&&& \multicolumn{4}{c}{None} & \multicolumn{4}{c}{Conv} & \multicolumn{4}{c}{Deconv} \\
        &P&U&C&\#params & mIoU & acc & bf & \#params & mIoU & acc & bf & \#params & mIoU & acc & bf \\
        \cline{1-4} \cline{5-8} \cline{9-12} \cline{13-16} \vspace{-1.0em} \\
        Linear              & 3 & - & 3 &   12.6M & 0.403 & 0.693 & 0.426                & \cellcolor{gray!25}25.1M & \cellcolor{gray!25}0.413 & \cellcolor{gray!25}0.699 & \cellcolor{gray!25}0.434       & 25.1M & 0.410 & 0.692 & 0.430 \\
        Standard Pool       & 2 & 3 & 3 &   0 & 0.352 & 0.657 & 0.375   & 12.6M & 0.377 & 0.675 & 0.393     & \cellcolor{gray!25}12.6M & \cellcolor{gray!25}0.397 & \cellcolor{gray!25}0.693 & \cellcolor{gray!25}0.407 \\
        \midrule
        MorphPool           & 2 & 3 & 3 &   0 & 0.385 & 0.682 & 0.441           & 12.6M & 0.399 & 0.688 & 0.449             & 12.6M & 0.410 & 0.694 & 0.454 \\
        MorphPool           & 3 & 5 & 3 &   0 & 0.425 & 0.707 &\bf0.476         & 12.6M & 0.428 & 0.714 & 0.479             & 12.6M & 0.442 & 0.720 & 0.487 \\
        Para. MorphPool     & 3 & 5 & 3 &   4.0K & \bf0.429 &\bf0.711 & 0.472   & 12.6M & 0.445 & 0.719 & 0.487             & 12.6M & 0.433 & 0.710 & 0.484 \\
        General MorphPool   & 3 & 5 & 3 &   67.5K & 0.425 & 0.710 & 0.473       & 12.6M & \bf0.445 & \bf0.722 & \bf0.488    & 12.6M & \bf0.450 & \bf0.724 & \bf0.492 \\
        \bottomrule
        \end{tabular}
    }
    \label{tab:full-segmentation-2d3ds-d}
\end{table*}

\begin{table*}
    \ra{1.05}
    \centering
    \caption{\textbf{SUN-RGBD Segmentation on Depth Results.}
    Results indicate morphological pooling and unpooling have improved performance over standard pooling and unpooling by a large margin.
    In addition, because of the non-linear nature of depth, morphological operations for down and up-sampling also beat the linear baseline with only half of the parameters.
    }
    \resizebox{1.0\textwidth}{!}{
        \begin{tabular}{@{\extracolsep{4pt}}lccc|cccc|cccc|cccc@{}}
        \toprule
        &&&& \multicolumn{4}{c}{None} & \multicolumn{4}{c}{Conv} & \multicolumn{4}{c}{Deconv} \\
        &P&U&C&\#params & mIoU & acc & bf & \#params & mIoU & acc & bf & \#params & mIoU & acc & bf \\
        \cline{1-4} \cline{5-8} \cline{9-12} \cline{13-16} \vspace{-1.0em} \\
        Linear              & 3 & - & 3 &   12.6M & 0.349 & 0.711 & 0.308     & \cellcolor{gray!25}25.1M & \cellcolor{gray!25}0.398 & \cellcolor{gray!25}0.741 & \cellcolor{gray!25}0.339      & 25.1M & 0.399 & 0.742 & 0.334 \\
        Standard Pool       & 2 & 3 & 3  &   0 & 0.208 & 0.643 & 0.255     & 12.6M & 0.297 & 0.687 & 0.294      & \cellcolor{gray!25}12.6M & \cellcolor{gray!25}0.288 & \cellcolor{gray!25}0.690 & \cellcolor{gray!25}0.294 \\
        \midrule
        MorphPool           & 2 & 3 & 3 &   0 & 0.345 & 0.712 & 0.328       & 12.6M & 0.375 & 0.735 & 0.345     & 12.6M & 0.375 & 0.732 & 0.347 \\
        MorphPool           & 3 & 5 & 3 &   0 & 0.382 & 0.735 & 0.346       & 12.6M & 0.412 & 0.748 & 0.364     & 12.6M & 0.407 & 0.747 & 0.361 \\
        Para. MorphPool     & 3 & 5 & 3 &  4.0K & 0.396 & 0.741 & 0.351    & 12.6M & 0.412 & \bf0.751 & 0.362  & 12.6M & 0.414 & \bf0.750 & \bf0.366 \\
        General MorphPool   & 3 & 5 & 3 &   67.5K & \bf0.398 & \bf0.745 & \bf0.352     & 12.6M & \bf0.416 & 0.748 & \bf0.366     & 12.6M & \bf0.416 & 0.748 & 0.365 \\
        \bottomrule
        \end{tabular}
    }
    \label{tab:full-segmentation-sun-d}
\end{table*}

\begin{table*}
    \ra{1.05}
    \centering
    \caption{\textbf{NYUv2 Segmentation on Depth Results.}
    Results on 2D-3D-S and SUN-RGBD are confirmed by the results on NYUv2.
    This is unsurprising: NYUv2 is a subset of the larger SUN-RGBD dataset.
    The quality of inpainting the NYUv2 depth maps is much better than other images in the SUN-RGBD dataset, since other subsets had much sparser depth data.
    }
    \resizebox{1.0\textwidth}{!}{
        \begin{tabular}{@{\extracolsep{4pt}}lccc|cccc|cccc|cccc@{}}
        \toprule
        &&&& \multicolumn{4}{c}{None} & \multicolumn{4}{c}{Conv} & \multicolumn{4}{c}{Deconv} \\
        &P&U&C&\#params & mIoU & acc & bf & \#params & mIoU & acc & bf & \#params & mIoU & acc & bf \\
        \cline{1-4} \cline{5-8} \cline{9-12} \cline{13-16} \vspace{-1.0em} \\
        Linear              & 3 & - & 3 &   12.6M & 0.305 & 0.597 & 0.175      & \cellcolor{gray!25}25.1M & \cellcolor{gray!25}0.320 & \cellcolor{gray!25}0.609 & \cellcolor{gray!25}0.176      & 25.1M & 0.312 & 0.603 & 0.176 \\
        Standard Pool       & 2 & 3 & 3  &   0 & 0.121 & 0.453 & 0.177     & 12.6M & 0.159 & 0.495 & 0.173      & \cellcolor{gray!25}12.6M & \cellcolor{gray!25}0.151 & \cellcolor{gray!25}0.487 & \cellcolor{gray!25}0.176 \\
        \midrule
        MorphPool           & 2 & 3 & 3 &   0 & 0.290 & 0.592 & 0.201    & 12.6M & 0.316 & 0.608 & 0.211        & 12.6M & 0.318 & 0.613 & 0.207 \\
        MorphPool           & 3 & 5 & 3 &   0 & 0.323 & 0.607 & 0.200    & 12.6M & 0.357 & 0.627 & 0.208        & 12.6M & 0.360 & 0.630 & 0.212 \\
        Para. MorphPool     & 3 & 5 & 3 &  4.0K & 0.348 & 0.627 & 0.205 & 12.6M & \bf0.360 & 0.630 & \bf0.212  & 12.6M & 0.367 & 0.630 & 0.215 \\
        General MorphPool   & 3 & 5 & 3 &   67.5K & \bf0.353 & \bf0.632 & \bf0.207      & 12.6M & 0.357 & \bf0.630 & 0.207      & 12.6M & \bf0.377 & \bf0.643 & \bf0.219 \\
        \bottomrule
        \end{tabular}
    }
    \label{tab:full-segmentation-nyu-d}
\end{table*}

\begin{table*}[]
    \ra{1.05}
    \centering
    \caption{\textbf{2D-3D-S Segmentation on RGB Results.}
    Similar to the segmentation results on depth, regular pooling and unpooling, even followed by (de)convolution, yields inferior performance to this paper's generalised morphological pooling.
    Linear interpolation, however, performs on par worse than morphological operations.
    This could be due to the RGB images, which cannot necessarily be expected to suit a set of morphological operations.
    Again, morphological operations introduce no or little additional parameters, whereas linear sampling plus convolution does.
    Finally, morphological pooling \& unpooling performs best at semantic boundaries as measured by the Boundary F1-score.
    }
    \resizebox{1.0\textwidth}{!}{
        \begin{tabular}{@{\extracolsep{4pt}}lccc|cccc|cccc|cccc@{}}
        \toprule
        &&&& \multicolumn{4}{c}{None} & \multicolumn{4}{c}{Conv} & \multicolumn{4}{c}{Deconv} \\
        &P&U&C&\#params & mIoU & acc & bf & \#params & mIoU & acc & bf & \#params & mIoU & acc & bf \\
        \cline{1-4} \cline{5-8} \cline{9-12} \cline{13-16} \vspace{-1.0em} \\
        Linear                  & 3 & - & 3 &   12.6M & 0.355 & 0.637 & 0.289       & \cellcolor{gray!25}25.1M & \cellcolor{gray!25}\bf0.370 & \cellcolor{gray!25}0.649 & \cellcolor{gray!25}0.301     & 25.1M & \bf0.372 & \bf0.650 & 0.302  \\
        Standard Pool           & 2 & 3 & 3  &   0 & 0.325 & 0.607 & 0.257    & 12.6M & 0.338 & 0.623 & 0.266     & \cellcolor{gray!25}12.6M & \cellcolor{gray!25}0.345 & \cellcolor{gray!25}0.625 & \cellcolor{gray!25}0.271 \\
        \midrule
        MorphPool               & 2 & 3 & 3 &   0 & 0.344 & 0.632 & 0.299     & 12.6M & 0.351 & 0.631 & 0.296     & 12.6M & 0.362 & 0.638 & 0.305 \\
        MorphPool               & 3 & 5 & 3 &   0 & \bf0.365 & \bf0.648 & 0.316     & 12.6M & 0.361 & 0.648 & 0.314      & 12.6M & 0.356 & 0.642 & \bf0.313 \\
        Para. MorphPool         & 3 & 5 & 3 &   4.0K & 0.362 & 0.642 & \bf0.317    & 12.6M & 0.354 & 0.642 & 0.308     & 12.6M & 0.365 & 0.649 & \bf0.313 \\
        General MorphPool       & 3 & 5 & 3 &   67.5K & 0.351 & 0.638 & 0.310     & 12.6M & 0.364 & \bf0.650 & \bf0.315      & 12.6M & 0.364 & 0.646 & 0.312 \\
        \bottomrule
        \end{tabular}
    }
    \label{tab:full-segmentation-2d3ds-i}
\end{table*}

\begin{table*}[]
    \ra{1.05}
    \centering
    \caption{\textbf{SUN-RGBD Segmentation on RGB Results.}
    SUN-RGBD shows similar results to 2D-3D-S on RGB input. 
    In this table, parameterising the structuring elements by either parabolic or general structuring elements is also shown.
    }
    \resizebox{1.0\textwidth}{!}{
        \begin{tabular}{@{\extracolsep{4pt}}lccc|cccc|cccc|cccc@{}}
        \toprule
        &&&& \multicolumn{4}{c}{None} & \multicolumn{4}{c}{Conv} & \multicolumn{4}{c}{Deconv} \\
        &P&U&C&\#params & mIoU & acc & bf & \#params & mIoU & acc & bf & \#params & mIoU & acc & bf \\
        \cline{1-4} \cline{5-8} \cline{9-12} \cline{13-16} \vspace{-1.0em} \\
        Linear              & 3 & - & 3 &   12.6M & 0.381 & 0.694 & 0.314      & \cellcolor{gray!25}25.1M & \cellcolor{gray!25}\bf0.396 & \cellcolor{gray!25}\bf0.707 & \cellcolor{gray!25}0.329      & 25.1M & \bf0.398 & \bf0.709 & 0.329 \\
        Standard Pool       & 2 & 3 & 3  &   0 & 0.298 & 0.646 & 0.270      & 12.6M & 0.351 & 0.682 & 0.312       & \cellcolor{gray!25}12.6M & \cellcolor{gray!25}0.353 & \cellcolor{gray!25}0.682 & \cellcolor{gray!25}0.310 \\
        \midrule
        MorphPool           & 2 & 3 & 3 &   0 & 0.348 & 0.675 & 0.330       & 12.6M & 0.368 & 0.697 & 0.346    & 12.6M & 0.381 & 0.699 & \bf0.349  \\
        MorphPool           & 3 & 5 & 3 &   0 & 0.387 & 0.695 & 0.334       & 12.6M & 0.363 & 0.701 & 0.342    & 12.6M & 0.379 & 0.705 & 0.344 \\
        Para. MorphPool     & 3 & 5 & 3 &  4.0K & 0.383 & 0.695 & 0.335     & 12.6M & 0.382 & 0.701 & 0.346    & 12.6M & 0.387 & 0.706 & \bf0.349 \\
        General MorphPool   & 3 & 5 & 3 &   67.5K & \bf0.388 & \bf0.698 & \bf0.337      & 12.6M & 0.394 & 0.704 & \bf0.352       & 12.6M & 0.390 & 0.707 & 0.346 \\
        \bottomrule
        \end{tabular}
    }
    \label{tab:full-segmentation-sun-i}
\end{table*}

\begin{table*}[]
    \ra{1.05}
    \centering
    \caption{\textbf{NYUv2 Segmentation on RGB Results.}
    Similar to 2D-3D-S and SUN-RGBD on image input, results show that \textbf{(a)} morphological pooling outperforms standard pooling; \textbf{(b)} morphological pooling performs on par or slightly better than linear sampling, although at much reduced parameter count; and \textbf{(c)} morphological operations perform better at semantic boundaries.
    }
    \resizebox{1.0\textwidth}{!}{
        \begin{tabular}{@{\extracolsep{4pt}}lccc|cccc|cccc|cccc@{}}
        \toprule
        &&&& \multicolumn{4}{c}{None} & \multicolumn{4}{c}{Conv} & \multicolumn{4}{c}{Deconv} \\
        &P&U&C&\#params & mIoU & acc & bf & \#params & mIoU & acc & bf & \#params & mIoU & acc & bf \\
        \cline{1-4} \cline{5-8} \cline{9-12} \cline{13-16} \vspace{-1.0em} \\
        Linear              & 3 & - & 3 &   12.6M & 0.321 & 0.578 & 0.191      & \cellcolor{gray!25}25.1M & \cellcolor{gray!25}\bf0.339 & \cellcolor{gray!25}0.595 & \cellcolor{gray!25}0.194      & 25.1M & \bf0.338 & \bf0.597 & 0.194 \\
        Standard Pool       & 2 & 3 & 3  &   0 & 0.193 & 0.514 & 0.198     & 12.6M & 0.281 & 0.555 & \bf0.234      & \cellcolor{gray!25}12.6M & \cellcolor{gray!25}0.271 & \cellcolor{gray!25}0.556 & \cellcolor{gray!25}\bf0.229 \\
        \midrule
        MorphPool           & 2 & 3 & 3 &   0 & 0.295 & 0.550 & 0.219       & 12.6M & 0.316 & 0.574 & 0.227     & 12.6M & 0.314 & 0.577 & 0.223  \\
        MorphPool           & 3 & 5 & 3 &   0 & 0.312 & 0.584 & 0.203       & 12.6M & 0.335 & \bf0.599 & 0.213     & 12.6M & 0.330 & 0.591 & 0.210 \\
        Para. MorphPool     & 3 & 5 & 3 &  4.0K & 0.309 & 0.571 & 0.193    & 12.6M & 0.328 & 0.588 & 0.209     & 12.6M & 0.325 & 0.582 & 0.206 \\
        General MorphPool   & 3 & 5 & 3 &   67.5K & \bf0.327 & \bf0.586 & \bf0.211      & 12.6M & 0.326 & 0.593 & 0.208      & 12.6M & 0.322 & 0.587 & 0.210 \\
        \bottomrule
        \end{tabular}
    }
    \label{tab:full-segmentation-nyu-i}
\end{table*}

\begin{table*}[t]
    \ra{1.05}
    \centering
    \caption{\textbf{NYUv2 Depth Auto-encoding Results.}
    Similar to semantic segmentation on depth, morphological operations are well suited to sampling features from a Depth modality.
    Morphological pooling \& unpooling outperform linear sampling and standard pooling on this task.
    Parameterising the morphological operations with a free kernel yields best performance.
    }
    \resizebox{1.0\textwidth}{!}{
        \begin{tabular}{@{\extracolsep{4pt}}lccc|ccc|ccc|ccc@{}}
        \toprule
        &&&& \multicolumn{3}{c}{None} & \multicolumn{3}{c}{Conv} & \multicolumn{3}{c}{Deconv} \\
        &P&U&C& ARD & RMS & acc $<$ 1.25 & ABS & RMS & acc$<$1.25 & ARD & RMS & acc$<$1.25 \\
        &&&   & $\downarrow$ & $\downarrow$ & $\uparrow$ & $\downarrow$ & $\downarrow$ & $\uparrow$ & $\downarrow$ & $\downarrow$ & $\uparrow$ \\
        \cline{1-4} \cline{5-7} \cline{8-10} \cline{11-13} \vspace{-1.0em} \\
        Linear              & 3 & - & 3 &   0.193 & 0.563 & 0.715     & \cellcolor{gray!25}0.190 & \cellcolor{gray!25}0.566 & \cellcolor{gray!25}0.716     & 0.199 & 0.594 & 0.703 \\
        Standard Pool       & 2 & 3 & 3 &   0.239 & 0.751 & 0.605      & 0.224 & 0.687 & 0.634     & \cellcolor{gray!25}0.225 & \cellcolor{gray!25}0.695 & \cellcolor{gray!25}0.633 \\
        \midrule
        MorphPool           & 2 & 3 & 3 &   0.180 & 0.526 & 0.733     & 0.166 & 0.496 & \bf0.774       & 0.180 & 0.523 & 0.738 \\
        MorphPool           & 3 & 5 & 3 &   0.177 &0.534 & 0.734     & 0.169 & 0.495 & 0.760     & 0.174 & \bf0.508 & 0.749 \\
        General MorphPool   & 3 & 5 & 3 &   \bf0.171 & \bf0.505 & \bf0.747      & \bf0.161 & \bf0.462 & \bf0.774      & \bf0.171 & 0.510 & \bf0.753 \\
        \bottomrule
        \end{tabular}
    }
    \label{tab:full-encoding-nyu}
\end{table*}

Additionally, depth-wise convolutions can be used to equalise the number of parameters available to networks making use of morphological and linear sampling. 
On top of the results on depth data for SUN-RGBD in the main text, results are listed for depth input on NYU in~\autoref{tab:full-segmentation-nyu-d-depthwise}, and for RGB input in~\autoref{tab:full-segmentation-sun-i-depthwise} and~\autoref{tab:full-segmentation-nyu-i-depthwise}.
In general, for networks with the same number of parameters MorphPool outperforms its linear counterpart.
Again, the effect is most pronounced for depth data.

\begin{table*}
    \ra{1.05}
    \centering
    \caption{\textbf{Depth-wise Convolution Results for Depth on NYU}. 
    Similar to in the main text, it is possible to use depth-wise convolutions to equalise the number of parameters available during up-sampling.
    Gray cells indicate networks that have the same number of available parameters to learn interpolation for up-sampling, either morphologically or linearly.
    Just like on SUN-RGBD, MorphPool outperforms the linear setting on NYU.
    }
    \resizebox{1.0\textwidth}{!}{
        \begin{tabular}{@{\extracolsep{4pt}}l|cccc|cccc|cccc@{}}
        \toprule
        \multirow{2}{*}{\diagbox[width=10em,trim=l]{Down}{Up}} & \multicolumn{4}{c}{None} & \multicolumn{4}{c}{Depth-wise Conv} & \multicolumn{4}{c}{Conv}\\
        & \#params & mIoU & acc & bf & \#params & mIoU & acc & bf & \#params & mIoU & acc & bf \\
        \cline{1-1} \cline{2-5} \cline{6-9} \cline{10-13} \vspace{-1.0em} \\
        Conv                    & 12576576 & 0.305 & 0.597 & 0.175    & 12626176 & 0.323 & 0.613 & 0.177     & 47494976 & 0.362 & 0.631 & 0.189 \\
        Depth-wise Conv         & 17856 & 0.283 & 0.583 & 0.163       & \cellcolor{gray!25}67456 & \cellcolor{gray!25}0.300 & \cellcolor{gray!25}0.596 & \cellcolor{gray!25}0.167         & 34936256 & 0.314 & 0.607 & 0.176 \\
        Standard Pool           & 0 & 0.121 & 0.453 & 0.177           & 49600 & 0.149 & 0.480 & 0.173        & 34918400 & 0.204 & 0.543 & 0.198 \\
        \midrule
        MorphPool               & 0 & 0.323 & 0.607 & 0.200     & 49600 & \bf0.367 & 0.637 & \bf0.211        & 34918400 & 0.361 & 0.627 & 0.213 \\
        Para. MorphPool         & 3968 & 0.348 & 0.627 & 0.205     & 53568 & 0.366 & \bf0.640 & 0.205        & 34922368 & 0.379 & 0.637 & 0.214 \\
        General MorphPool       & \cellcolor{gray!25}67456 & \cellcolor{gray!25}\bf0.353 & \cellcolor{gray!25}\bf0.632 & \cellcolor{gray!25}\bf0.207       & 117056 & 0.359 & 0.633 & 0.205       & 34985856 & \bf0.380 & \bf0.643 & \bf0.217 \\
        \bottomrule
        \end{tabular}
    }
    \label{tab:full-segmentation-nyu-d-depthwise}
\end{table*}
\begin{table*}
    \ra{1.05}
    \centering
    \caption{\textbf{Depth-wise Convolution Results for RGB on SUN-RGB}. 
    Similar to the experiments that compare depth and RGB input, the difference in performance between pooling methods is less pronounced. 
    However, MorphPool outperforms the linear network with the same number of parameters.
    }
    \resizebox{1.0\textwidth}{!}{
        \begin{tabular}{@{\extracolsep{4pt}}l|cccc|cccc|cccc@{}}
        \toprule
        \multirow{2}{*}{\diagbox[width=10em,trim=l]{Down}{Up}} & \multicolumn{4}{c}{None} & \multicolumn{4}{c}{Depth-wise Conv} & \multicolumn{4}{c}{Conv}\\
        & \#params & mIoU & acc & bf & \#params & mIoU & acc & bf & \#params & mIoU & acc & bf \\
        \cline{1-1} \cline{2-5} \cline{6-9} \cline{10-13} \vspace{-1.0em} \\
        Linear                  & 12576576 & 0.381 & 0.694 & 0.314    & 12626176 & 0.418 & \bf0.726 & 0.346     & 47494976 & 0.353 & \bf0.717 & 0.325 \\
        Depth-wise Linear       & 17856 & 0.368 & 0.684 & 0.301       & \cellcolor{gray!25}67456 & \cellcolor{gray!25}0.383 & \cellcolor{gray!25}0.706 & \cellcolor{gray!25}0.326     & 34936256 & 0.393 & 0.705 & 0.321 \\
        Standard Pool           & 0 & 0.298 & 0.646 & 0.270           & 49600 & 0.376 & 0.696 & 0.319        & 34918400 & 0.382 & 0.701 & 0.333 \\
        \midrule
        MorphPool               & 0 & 0.387 & 0.695 & 0.334           & 49600 & 0.418 & 0.720 & 0.364        & 34918400 & 0.395 & 0.714 & \bf0.354 \\
        Para. MorphPool         & 3968 & 0.383 & 0.695 & 0.335        & 53568 & 0.417 & 0.722 & 0.364        & 34922368 & \bf0.397 & 0.715 & 0.353 \\
        General MorphPool       & \cellcolor{gray!25}67456 & \cellcolor{gray!25}\bf0.388 & \cellcolor{gray!25}\bf0.698 & \cellcolor{gray!25}\bf0.337       & 117056 & \bf0.419 & 0.723 & \bf0.365       & 34985856 & 0.392 & 0.713 & 0.348 \\
        \bottomrule
        \end{tabular}
    }
    \label{tab:full-segmentation-sun-i-depthwise}
\end{table*}
\begin{table*}[t!]
    \ra{1.05}
    \centering
    \caption{\textbf{Depth-wise Convolution Results for RGB on NYU}. 
    Again, MorphPool outperforms the linear setting at the same number of parameters.
    }
    \resizebox{1.0\textwidth}{!}{
        \begin{tabular}{@{\extracolsep{4pt}}l|cccc|cccc|cccc@{}}
        \toprule
        \multirow{2}{*}{\diagbox[width=10em,trim=l]{Down}{Up}} & \multicolumn{4}{c}{None} & \multicolumn{4}{c}{Depth-wise Conv} & \multicolumn{4}{c}{Conv}\\
        & \#params & mIoU & acc & bf & \#params & mIoU & acc & bf & \#params & mIoU & acc & bf \\
        \cline{1-1} \cline{2-5} \cline{6-9} \cline{10-13} \vspace{-1.0em} \\
        Conv                    & 12576576 & 0.321 & 0.578 & 0.191    & 12626176 & 0.344 & 0.519 & 0.201   & 47494976 & \bf0.351 & \bf0.607 & 0.196 \\
        Depth-wise Conv         & 17856 & 0.299 & 0.561 & 0.180       & \cellcolor{gray!25}67456 &  \cellcolor{gray!25}0.310 &  \cellcolor{gray!25}0.568 &  \cellcolor{gray!25}0.189       & 34936256 & 0.322 & 0.579 & 0.181 \\
        Standard Pool           & 0 & 0.193 & 0.514 & 0.198           & 49600 & 0.275 & 0.565 & 0.222        & 34918400 & 0.310 & 0.577 & \bf0.236 \\
        \midrule
        MorphPool               & 0 & 0.312 & 0.584 & 0.203           & 49600 & 0.332 & 0.594 & 0.227         & 34918400 & 0.317 & 0.583 & 0.192 \\
        Para. MorphPool         & 3968 & 0.309 & 0.571 & 0.193        & 53568 & 0.333 & 0.594 & 0.226        & 34922368 & 0.331 & 0.588 & 0.203 \\
        General MorphPool       & \cellcolor{gray!25}67456 &  \cellcolor{gray!25}\bf0.327 &  \cellcolor{gray!25}\bf0.586 &  \cellcolor{gray!25}\bf0.211       & 117056 & \bf0.344 & \bf0.604 & \bf0.232        & 34985856 & 0.336 & 0.594 & 0.207 \\
        \bottomrule
        \end{tabular}
    }
    \label{tab:full-segmentation-nyu-i-depthwise}
\end{table*}
\fi

\end{document}